\documentclass[journal]{IEEEtran}
\usepackage{amsmath}
\usepackage{graphicx}
\usepackage{epstopdf}
\usepackage{subfigure}
\usepackage{extarrows}
\usepackage{bm}
\usepackage{amssymb}
\usepackage{multirow}

\ifCLASSINFOpdf
\else
\fi
\begin{document}
%
\title{Sar Image Despeckling Based on Nonlocal Similarity Sparse Decomposition}
%
%
%

\author{Cheng-wei~Sang,~
        Hong~Sun,~\IEEEmembership{Member,~IEEE}
\thanks{This work was supported in part by the National Natural
Science Foundation of China under Grants 60872131  and F011202.}
\thanks{Cheng-wei Sang is with School of Electronic
Information, Wuhan University, Wuhan 430072, China (e-mail: SangCW@whu.
edu.cn).}
\thanks{Hong Sun is with School of Electronic
Information, Wuhan University, Wuhan 430072, China and Institut Telecom ParisTech, 46 rue Barrault, 75634 Paris, France (e-mail: hongsun@whu.edu.cn).}
\thanks{}}

%
%

\markboth{}%
{Shell \MakeLowercase{\textit{et al.}}: Bare Demo of IEEEtran.cls for Journals}
%



\maketitle

\begin{abstract}
This letter presents a method of synthetic aperture radar (SAR) image despeckling aimed to preserve the detail information while suppressing speckle noise. This method combines the nonlocal self-similarity partition and a proposed modified sparse decomposition. The nonlocal partition method groups a series of structure-similarity data sets. Each data set has a good sparsity for learning an over-complete dictionary in sparse representation. In the sparse decomposition, we propose a novel method to identify principal atoms from over-complete dictionary to form a principal dictionary. Despeckling is performed on each data set over the principal dictionary with principal atoms. Experimental results demonstrate that the proposed method can achieve high performances in terms of both speckle noise reduction and structure details preservation.
\end{abstract}


%
\IEEEpeerreviewmaketitle

\section{Introduction}
%
%
%
%
\IEEEPARstart{S}{peckle} noise is an inherent property of SAR imaging system. This speckle phenomenon not only degrades the image legibility but also affects target detection, segmentation, edge detection and so on. Despeckling is, therefore, an important preprocessing for the application of SAR image. Many classical methods denoise SAR images in spatial domain, such as \cite{IEEE:one}--\cite{IEEE:three}. It has been proven that denoising in transform domain, instead of denoising in spatial domain, can achieve better results, such as wavelet shrinkages \cite{IEEE:four}--\cite{IEEE:seven}. However, the fixed wavelet bases are not able to provide an adequate representation of various image structures (edges and textures). In fact, a better tradeoff between speckle smoothing and details preservation is always a challenge.
 \newline
\indent
The nonlocal means algorithm (NLM) \cite{IEEE:eight} provides a potential breakthrough to solve the natural image denoising problem. NLM takes advantage of nonlocal self-similarity embedded in natural images to achieve excellent performance. In fact, nonlocal similarity is also an inherent property of SAR images \cite{IEEE:seven}. Inspired by NLM, nonlocal-based methods \cite{IEEE:nine}--\cite{IEEE:eightteen} have been extended to SAR image despeckling. A representative work is the probabilistic patch-based (PPB) algorithm \cite{IEEE:eleven}, which selects homogeneous pixels by self-similarity and averages out the noise among homogeneous pixels. The nonlocal self-similarity helps PPB to effectively suppress speckle noise, but estimation by weighted average often makes reduction of fine detail information.
 \newline
\indent
The sparse decomposition method has recently become a strong tool in image denoising. It has shown in \cite{IEEE:twelve}--\cite{IEEE:thirteen} that the over-complete dictionary learned from natural images gets better capability of representing structural features (e.g., image edges and texture) and leads to better results than wavelet shrinkages. This success is mainly due to the fact that natural images can be sparsely represented by a few bases. Sparse decomposition has been extended to despeckling \cite{IEEE:fourteen}--\cite{IEEE:sixteen}. But the sparsity of SAR image is not enough for sparse representation. In addition, the noise level of SAR image is too high to be suppressed by an over-complete dictionary.
 \newline
\indent
In this letter, we propose a novel despeckling method by a new sparse decomposition method combined by the nonlocal self-similarity partition. The sparsity of SAR data is enhanced by nonlocal self-similarity partition, so that even finest details embedded in groups of similarity data sets can be sparsely represented. The sparse decomposition uses signal principal dictionary identified from over-complete dictionary, so that the strong noise can be suppressed.
 \newline
\indent
This letter is organized as follows. Section II details the proposed algorithm. In Section III, we show the experimental results. Section IV concludes this letter.

\section{Proposed Method}
We proposed a novel SAR image despeckling method based on a principal sparse decomposition with the nonlocal self-similarity.
\subsection{The Despeckling Framework}
The proposed despeckling method mainly includes two basis steps: sparsity-driven partition and principal dictionary denoising (PDD). Fig. 1 illustrates a block diagram of our method. Concretely, first, sparsity-driven partition step stacks structure similarity image patches into groups by patch matching. The true signals in such group of patches are mostly homogeneous. Thus the signals should have a sparse expansion over the learned over-complete dictionary, and achieve high degree of sparsity. Sparsity-driven partition enhances the sparsity of SAR images, so that the details underlying in similarity image patches can be sparsely represented.
 \newline
\indent
Second, the PDD learns over-complete dictionary from structure similarity image patches. The learned over-complete dictionary adequately reflects the intrinsic structure characteristic of image patches, so that the details can be well preserved. However, the performance of the learned over-complete dictionary degrades when training data set are contaminated by strong noise \cite{IEEE:seventeen}. It is because that the strong noise get into the over-complete dictionary. Since noise level of SAR image is very high, noise incursion makes speckle noise cannot be well suppressed by an over-complete dictionary. The PPD identifies signal principal atoms from over-complete dictionary to form the principal dictionary. We thus make benefit from learned over-complete dictionary which preserves details of image and from the principal dictionary which rejects strong noise. Then, despeckling is performed on the structure similarity image patches over such principal dictionary. A better balance between suppressing noise and preserving details can be achieved by PPD.
\begin{figure*}[htbp]
  \begin{center}
  \includegraphics[width=0.62 \linewidth]{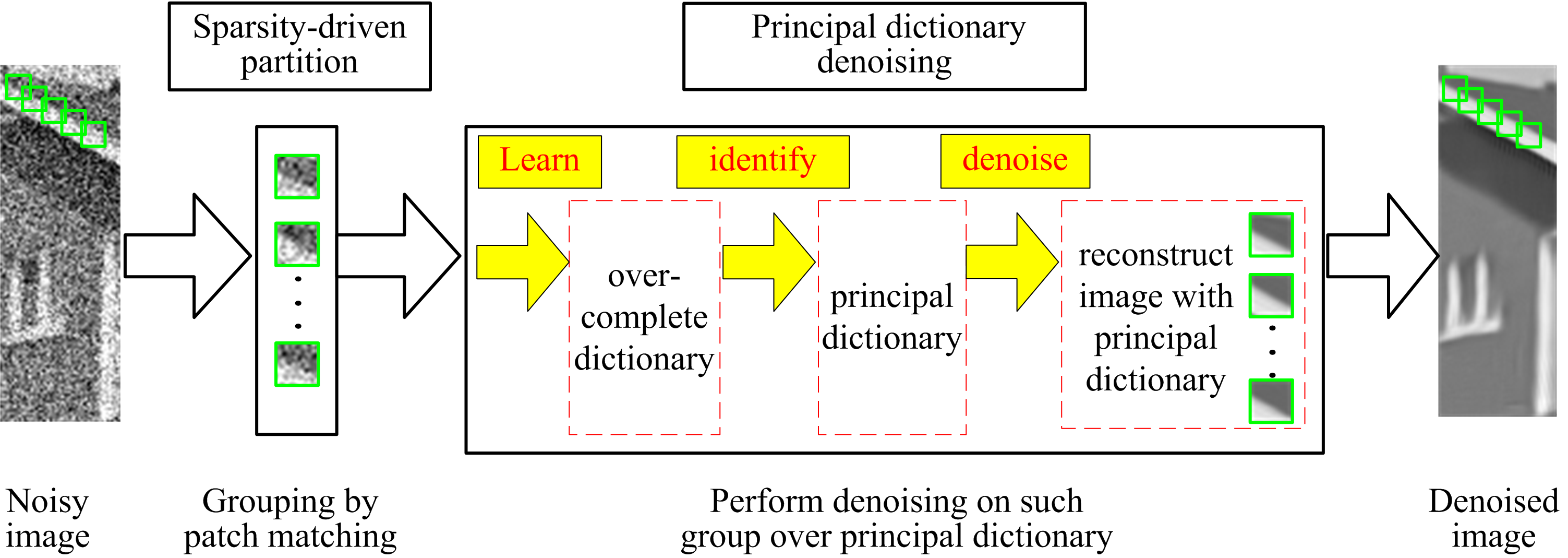}
  \caption{Block diagram of the proposed approach}\label{figure1}
  \end{center}
\end{figure*}
\newline
\indent
Under the assumption that the speckle noise is fully developed, SAR image can be expressed as
\begin{equation}\label{2}
\begin{aligned}
   I=Su
\end{aligned}
\end{equation}
where $I$, $S$ and $u$ represent observed intensity image, noise-free reflectance and speckle noise, respectively.
\newline
\indent
Since the sparse decomposition model works under the assumption of additive noise model, the multiplicative model needs to be transformed to additive model by logarithmic transformation \cite{IEEE:eightteen} as follow
\begin{equation}\label{2}
\begin{aligned}
   \mathbf{Z}= \mathbf{X} + \mathbf{V}
\end{aligned}
\end{equation}
where $\mathbf{Z}=\log(I)$, $\mathbf{X}=\log(S)$, and $\mathbf{V}=\log(u)$. Noise $\mathbf{V}$ can be approximately assumed to be additive white noise.
\subsection{Sparsity-Driven Partition}
Consider an observation image of size $\sqrt{S} \times \sqrt{S}$ pixels, lexicographically ordered as column vector $\mathbf{Z} \in R^S$. Denote  $\mathbf{z}_i \in R^N$ as the vector representations of an image patch with size $\sqrt{N} \times \sqrt{N}$ at location $i$. Every patch $\mathbf{z}_i = \mathbf{R}_i \mathbf{Z}, \mathbf{R}_i \in R^{N \times S}$ is a matrix operator that extracts path $\mathbf{z}_i$ from $\mathbf{Z}$. In this letter we extract overlapped patches, and such patch-based representation is highly redundant representation.
\newline
\indent
SAR images with multiplicative noise are very different from the natural images, the Euclidean distance as a classical similar measure under the assumption of Gaussian noise is not optimal anymore. Considering multiplicative noise,the patch-based similarity measure proposed in PPB \cite{IEEE:eleven} is well suitable for SAR images to measure the structure similarity between patches $\mathbf{z}_i$ and $\mathbf{z}_j$ as follows:
\begin{equation}\label{2}
\begin{aligned}
   d_{(\mathbf{z}_i,\mathbf{z}_j)}=(2L-1){\sum\nolimits_{k}\log( \sqrt{\frac{\mathbf{A}_i(k)}{\mathbf{A}_j(k)}} + \sqrt{\frac{\mathbf{A}_j(k)}{\mathbf{A}_i(k)}}  )}
\end{aligned}
\end{equation}
where $\mathbf{A}_i=exp(\mathbf{z}_i)$, $exp(\cdot)$ is exponential transformation, and $L$ is the equivalent number of looks (ENL). The smaller $d_{(\mathbf{z}_i,\mathbf{z}_j)}$ is, the more similar $\mathbf{z}_i$ and $\mathbf{z}_j$ is.
\newline
\indent
For a given reference patch, this step aims to find $M$-most similar patches by similarity measure (3) in a large search area and stack them into a group $\mathbf{Z}_{\Omega_i}$ as follows:
\begin{equation}\label{2}
\begin{aligned}
    \mathbf{Z}_{\Omega_i}=\left\{\mathbf{z}_k\right\}_{k \in \Omega_i}
    \end{aligned}
\end{equation}
where
\begin{equation*}\label{2}
\begin{aligned}
\Omega_i \triangleq \{j=1,\cdots,M \ s.t. \ d_{(\mathbf{z}_i,\mathbf{z}_1)} \leq d_{(\mathbf{z}_i,\mathbf{z}_j)} \cdots \leq d_{(\mathbf{z}_i,\mathbf{z}_M)} \}
    \end{aligned}
\end{equation*}
\subsection{Proposed Principal Dictionary Denoising}
Sparse representation is commonly used image denoising approach. It assumes that an observation image can be well reconstructed by a linear combination of few components. For each group $\mathbf{Z}_{\Omega} \in \mathbf{R}^{N \times M}$ (4), the sparse representation model is defined by
\begin{equation}\label{2}
\begin{aligned}
    \operatorname*{arg min}\limits_{\mathbf{D},{\bm \alpha_m}} \parallel {\bm \alpha_m} \parallel_0 s.t. {\sum\nolimits_{m=1}^{M} \parallel \mathbf{D}{\bm \alpha}_m - \mathbf{x}_m\parallel^2_2} \leq T
\end{aligned}
\end{equation}
where $\parallel \cdot \parallel_0$ is $\ell^0$-norms, $\mathbf{z}_m \in R^{N \times 1}$ is an image patch of groups $\mathbf{Z}_{\Omega}$, $\bm{\alpha}_m \in R^{K \times 1} $ is sparse representation of $\mathbf{z}_m$, $\mathbf{D}=\{\mathbf{d}_k\}_{k=1}^{K} \in R^{N\times M}(K > M)$ is an over-complete dictionary, $\mathbf{d}_k$ is an atom of dictionary $\mathbf{D}$ , and $T$ is a bounded representation error.
\newline
\indent
The dictionary $\mathbf{D}$ can be obtained by K-SVD \cite{IEEE:twelve}, and the sparse coefficient matrix $\mathbf{A}_{K \times M} = \{ \bm{\alpha}_m \}_{m=1}^M$ can be solved by orthogonal matching pursuit algorithm. When we obtain dictionary $\mathbf{D}$ and and coefficient matrix $\mathbf{A}$, the estimate of underlying signals embedded in $\mathbf{Z}_{\Omega}$ is reconstructed by
\begin{equation}\label{2}
\begin{aligned}
   {\mathbf{ \hat {X}}_{\Omega}}= \mathbf{DA}
    \end{aligned}
\end{equation}
Then, all denoised groups ${\mathbf{\hat {X}}_{\Omega}}$s  are aggregated to form the denoised image $\hat{\mathbf{X}}$.
\newline
\indent
We note that the dictionary $\mathbf{D}$ is trained on noisy images $\mathbf{Z}_{\Omega}$, the strong noise inevitable get into the over-complete dictionary. So the atoms $\{\mathbf{d}\}_{k=1}^K$ are better to be subdivided into signal principal dictionary relating to signal and residual dictionary relating to noise residual. However, it is impossible to exploit an energy-constrained division since atoms $\{\mathbf{d}\}_{k=1}^K$ are not necessarily orthogonal or independent.
\newline
\indent
Let us consider coefficient matrix $\mathbf{A}$ (6) in terms of its row vectors, $\bm{\lambda}_k \in R^{1\times M}$
 \begin{equation}\label{Eq_6}
\begin{aligned}
 \mathbf{A}& = \left[\bm{\alpha}_1 \; \bm{\alpha}_2 \; \cdots \; \bm{\alpha}_M \right]\\& = \begin{bmatrix} \alpha_1(1) \quad \alpha_2(1) \quad \cdots \quad \alpha_M(1) \\ \alpha_1(2) \quad \alpha_2(2) \quad \cdots \quad \alpha_M(2)\\\vdots \quad \quad \quad \vdots \quad \quad \ddots \quad \quad \vdots \\ \alpha_1(K) \quad \alpha_2(K) \quad \cdots \quad \alpha_M(K)   \end{bmatrix} =  \begin{bmatrix} \bm{\lambda}_1 \\ \bm{\lambda}_2 \\ \vdots \\ \bm{\lambda}_K \end{bmatrix}  \\&
 \mbox{where} \quad \bm{\lambda}_k=\left[ \alpha_1(k)  \; \alpha_2(k) \; \dots \; \alpha_M(k) \right] \in R^{1 \times M}
   \end{aligned}
\end{equation}
Thus (7) becomes
 \begin{equation}\label{Eq_7}
\begin{aligned}
   \hat{\mathbf{X}}_{\Omega}& = \mathbf{D} \mathbf{A} \\& = \left[\mathbf{d}_1 \cdots \mathbf{d}_k \cdots \mathbf{d}_K \right] . \left[\bm{\lambda}_1^T \cdots \bm{\lambda}_k^T \cdots \bm{\lambda}_K^T \right]^T
   \end{aligned}
\end{equation}
Note that $\bm{\lambda}_k$ is not necessarily sparse vector.
\newline
\indent
Eqn. (8) tells us that row vector $\bm{\lambda}_k$ is the weight of the atom $\mathbf{d}_k$, which is a global parameter over the group $\mathbf{Z}_{\Omega}$.The $\parallel \bm{\lambda}_k \parallel_0$ is the number of occurrences of atom $\mathbf{d}_k$ over the group $\mathbf{Z}_{\Omega}$. We call it the frequency of the atom $\mathbf{d}_k$ denoted by $f_k$
\begin{equation}\label{Eq_8}
\begin{aligned}
   f_k\triangleq Frequency(\mathbf{d}_k | \mathbf{Z}_{\Omega}) = \| \bm{\lambda}_k \|_0
   \end{aligned}
\end{equation}
\newline
\indent
In sparse representation model, atoms $\{\mathbf{d}_k\}_{k=1}^K$ are prototype of signal patterns. That allows us consider them as a signal patterns. Thus, some essential features of this signal pattern could be considered as a criterion to identify important atoms. Considering the signals embedded in group $\mathbf{Z}_{\Omega}$ are mostly homogeneous, the underlying intrinsic features represented by principal bases must occur in group $\mathbf{Z}_{\Omega}$ with high frequency even with a lower energy, on the contrary, any noise pattern does not have structural redundancies in observed data even with a high energy and thus could not repeat. Intuitively, it seems that frequency $f_k$ is an important characteristic of intrinsic features of group $\mathbf{Z}_{\Omega}$. In addition, it has been proven \cite{IEEE:nineteen} that $f_k$ is a good description of the signal texture.
\newline
\indent
Thus this frequency $f_k$ is a reasonable criterion to identify principal atoms from over-complete dictionary to form principal dictionary.
\newline
\indent
PDD seeks atoms with high frequency to construct principal dictionary. The vectors $\{\bm{\lambda}_k\}_{k=1}^K$ are ranked by descending order of their frequency $f_k$ as
\begin{equation}\label{Eq_8}
\begin{aligned}
\Lambda &\triangleq  \left[\bm\lambda'_1\cdots\bm\lambda'_k\cdots\bm\lambda'_K \right] \xLongleftarrow{sort}  \left[\bm\lambda_1\cdots\bm\lambda_k\cdots\bm\lambda_K \right] \\&
where \parallel\bm\lambda'_1\parallel_0 \geq \parallel\bm\lambda'_2\parallel_0 \geq \cdots \geq \parallel\bm\lambda'_K\parallel_0
\end{aligned}
\end{equation}
With the reordered atoms $\{\mathbf{d}_k\}_{k=1}^K$ according to $\{\parallel\bm\lambda'_k \parallel_0\}^K_{k=1}$, the reordered dictionary $\mathbf{D}'$ is rewritten as:
\begin{equation}\label{Eq_9}
 \mathbf{D}'=\left[\mathbf{d}'_1 \cdots\mathbf{d}'_k \cdots\mathbf{d}'_K \right]
\end{equation}
Then
\begin{equation}\label{Eq_9}
 \mathbf{X}_{\Omega}=\mathbf{DA}=\mathbf{D'}\Lambda
\end{equation}
We take the first $P(P<K)$ atoms to form a principal dictionary $\mathbf{\bar{D}}_P^{\mathbf{S}}$ as
\begin{equation}\label{Eq_9}
\begin{aligned}
  \mathbf{\bar{D}}_{P}^{\mathbf{S}} = \left[\mathbf{d}'_1 \quad \mathbf{d}'_2 \cdots \mathbf{d}'_P   \right]
  \end{aligned}
\end{equation}
 The remaining atoms are truncated as residual dictionary $\mathbf{\bar{D}}_{K-P}^{\mathbf{R}}$ as
\begin{equation}\label{Eq_9}
\begin{aligned}
  \mathbf{D}_{K-P}^{\mathbf{R}}=\left[\mathbf{d}'_{P+1} \quad \mathbf{d}'_{P+2} \cdots \mathbf{d}'_K   \right]
  \end{aligned}
\end{equation}
\newline
\indent
The threshold $P$ of the image base (13) could be simply decided according to the histogram of $\{\parallel\bm\lambda'_k \parallel_0\}^K_{k=1}$.  One can set maximum point of the histogram to $P$ as
\begin{equation}\label{Eq_11}
P= \operatorname*{arg max}\limits_{k} Hist\left( \parallel\bm\lambda'_{k}\parallel_0 \right)
\end{equation}
\newline
\indent
We recover signals $\mathbf{\hat{X}}_{\Omega}^{\mathbf{S}}$ from $\mathbf{Z}_{\Omega}$ with principal dictionary $\mathbf{\bar{D}}_P^{\mathbf{S}}$ as
\begin{equation}\label{Eq_9}
\begin{aligned}
 \mathbf{\hat{X}}_{\Omega}^{\mathbf{S}} &= \mathbf{\bar{D}}_P^{\mathbf{S}} \Lambda_P^{\mathbf{S}} \\&=\left[\mathbf{d}'_{1} \quad \mathbf{d}'_{2} \cdots \mathbf{d}'_K   \right] [{\bm\lambda'}_{1}^{T} \quad {\bm\lambda'}_{2}^T \cdots {\bm\lambda'}_{P}^T ]^T
  \end{aligned}
\end{equation}
\newline
\indent
Because the patches are overlapping, each pixel can be included in many groups, with more than one estimated values. Thus these estimated values of each pixel in every denoised group $\mathbf{\hat{X}}_{\Omega}^{\mathbf{S}}$ need to be aggregated to form denoised image $\mathbf{\hat{X}}^{\mathbf{S}}$ with weighted average. We show an example of the PDD in Fig. 2.
\begin{figure}[htbp]
  \begin{center}
  \includegraphics[width=0.90 \linewidth]{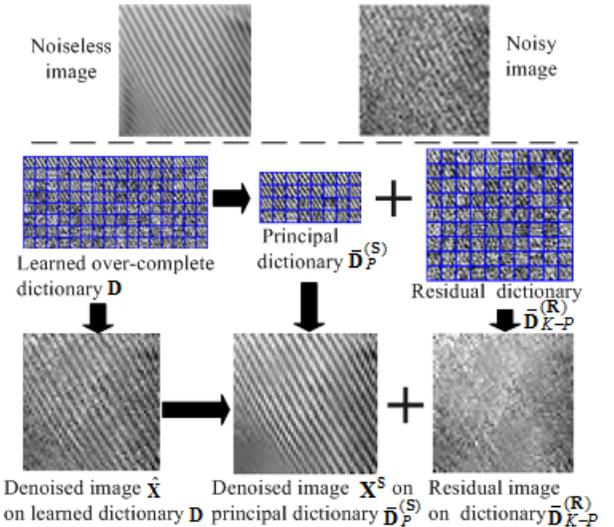}
  \caption{Example of principal dictionary denoising}\label{figure1}
  \end{center}
\end{figure}
\newline
\indent
Due to the log-transformation is used to transform SAR images to additive noise model, we need to correct nonzero mean as proposed by \cite{IEEE:six} before PDD and take the exponential transformation of $\mathbf{\hat{X}}^{\mathbf{S}}$ to produce the final denoised image.

\section{Experimental Results}
To demonstrate the efficacy of the proposed despeckling algorithm, in this section, we give our experimental results including both simulated and real SAR images shown in Fig. 3. Since SAR image is lack of the original noise-free image, in order to achieve a quantitative evaluation, optical image degraded by simulated speckle noise is used to test the despeckling performance for $L=1,2,4$ and $8$ looks. Thus, we can treat the original optical image as the noise-free image and do soundly comparison with other denoising algorithms. For the real SAR image, we discuss experiments based on visual interpretation.
\begin{figure}[htbp]
\begin{center} 
\subfigure[]{
\includegraphics[width=0.4 \linewidth]{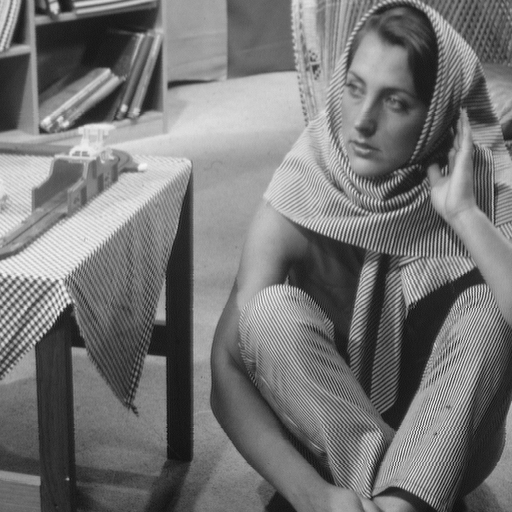}}
\centering
\subfigure[]{
\includegraphics[width=0.4 \linewidth]{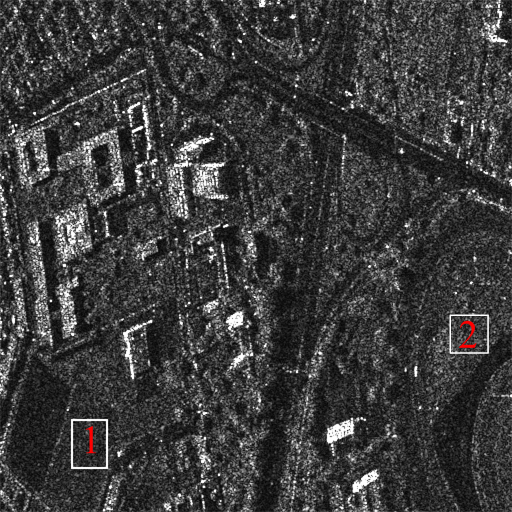}}
\caption{Images used in the experiments. (a) Barbara ($512 \times 512$) (b) TerraSAR-X sample image ($512 \times 512$), $L=1$.}
\end{center}
\end{figure}
\begin{figure}[bht]
 \begin{minipage}[t]{0.49\linewidth}
  \centerline{  \includegraphics[width=0.99 \linewidth]{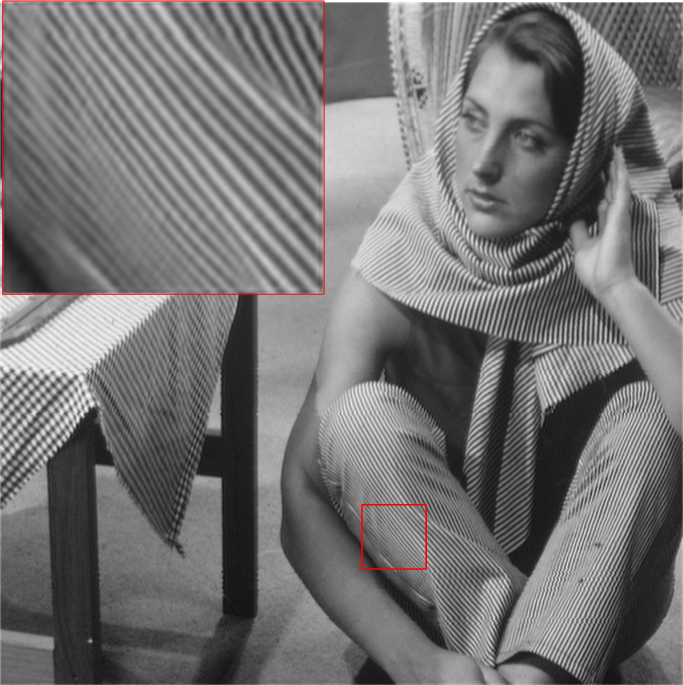} }
  \centerline{(a)}
 \end{minipage}
 \hfill
  \begin{minipage}[t]{0.49\linewidth}
  \centerline{  \includegraphics[width=0.99 \linewidth]{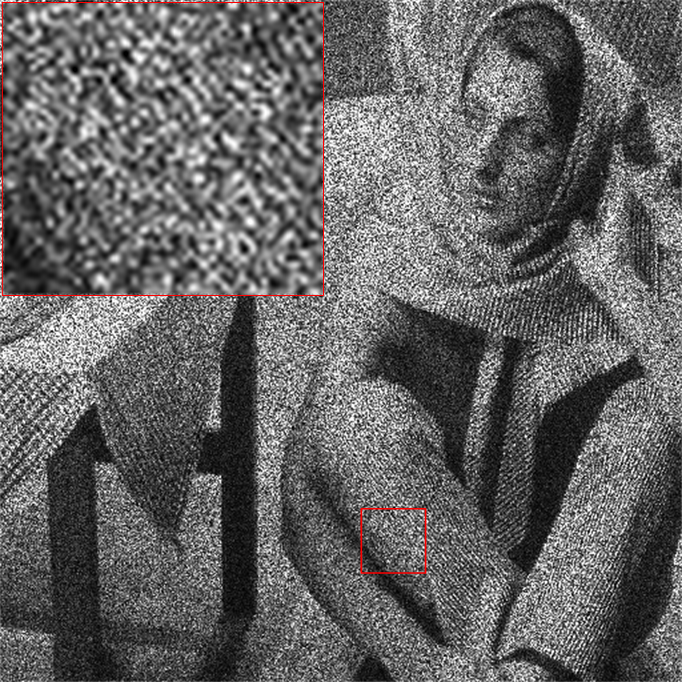} }
  \centerline{(b)}
 \end{minipage}
 \vfill
  \begin{minipage}[t]{0.48\linewidth}
  \centerline{  \includegraphics[width=0.99 \linewidth]{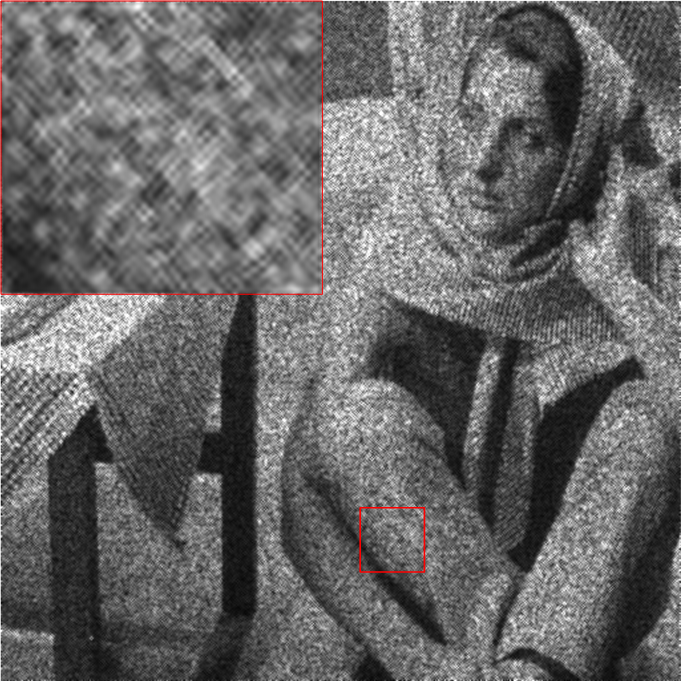} }
  \centerline{(c)}
 \end{minipage}
 \hfill
  \begin{minipage}[t]{0.49\linewidth}
  \centerline{  \includegraphics[width=0.99 \linewidth]{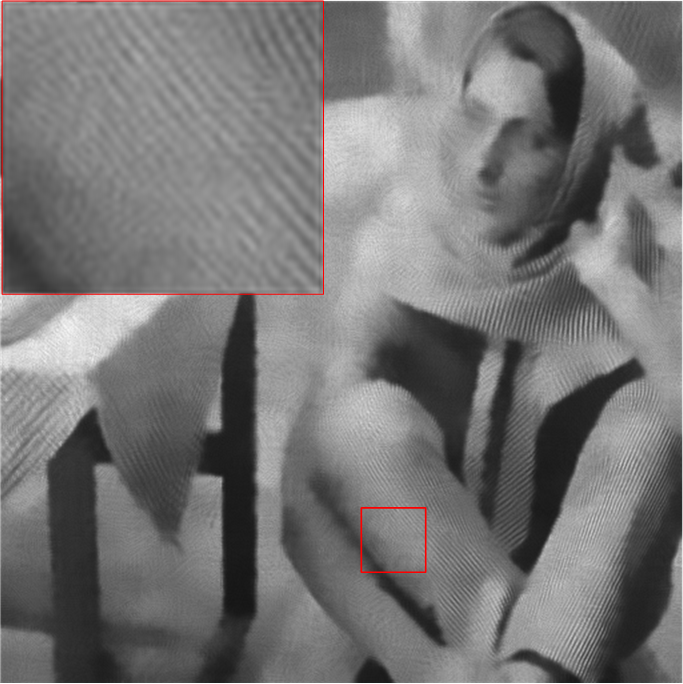} }
  \centerline{(d)}
 \end{minipage}
  \vfill
  \begin{minipage}[t]{0.49\linewidth}
  \centerline{  \includegraphics[width=0.99 \linewidth]{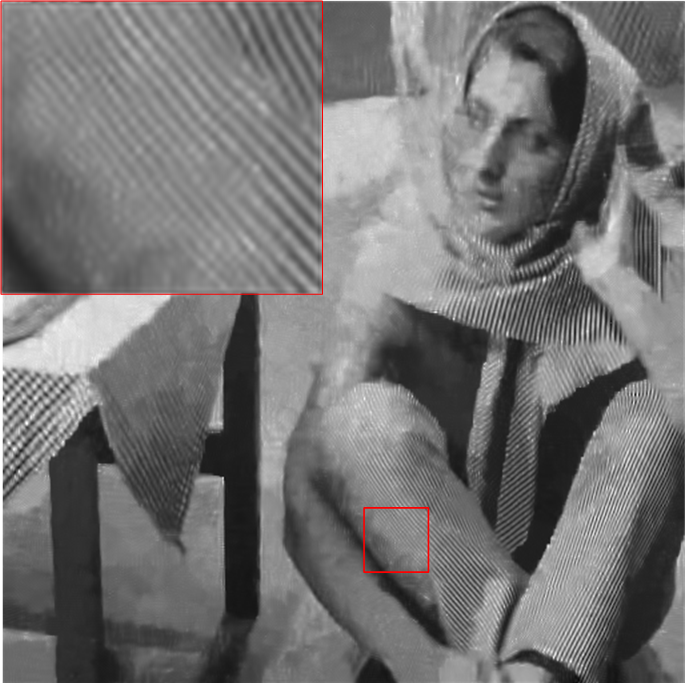} }
  \centerline{(e)}
 \end{minipage}
 \hfill
  \begin{minipage}[t]{0.49\linewidth}
  \centerline{  \includegraphics[width=0.99 \linewidth]{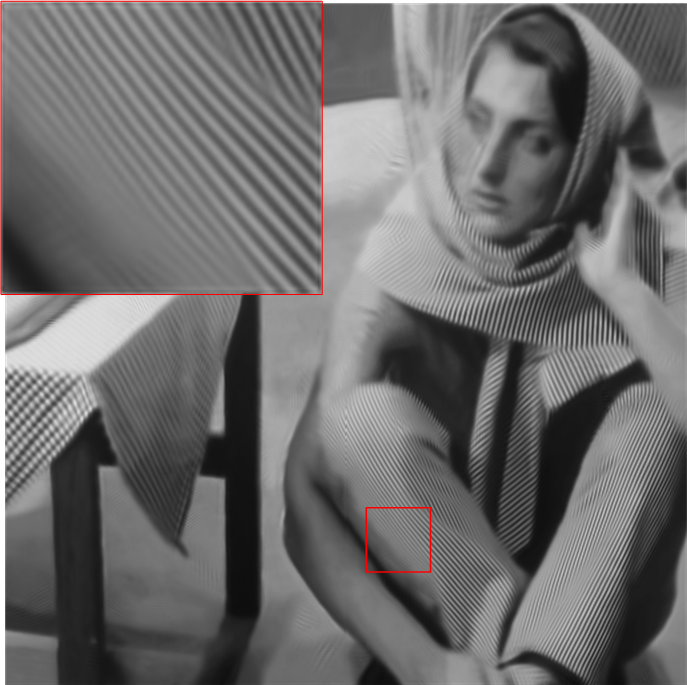} }
  \centerline{(f)}
 \end{minipage}
  \caption{Zoom of filtered images for Barbara corrupted by one-look speckle. (a) Noise-free Barbara (b) Noisy image. (c) Frost. (d) PPB. (e) SAR-BM3D. (f) Proposed method.}\label{figure4}
\end{figure}
%

\begin{figure*}[htbp]
  \centering
  \subfigure[]{\includegraphics[width=0.22 \linewidth]{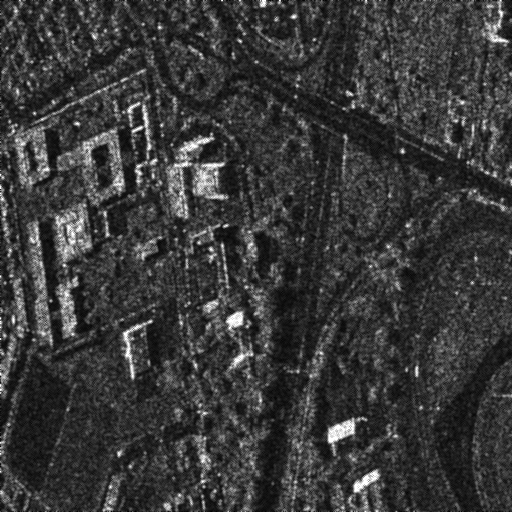} \label{a1}}
  \subfigure[]{\includegraphics[width=0.22 \linewidth]{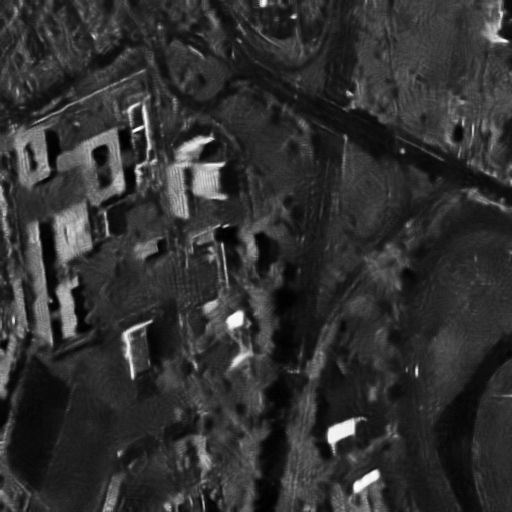} \label{s3}}
  \subfigure[]{\includegraphics[width=0.22 \linewidth]{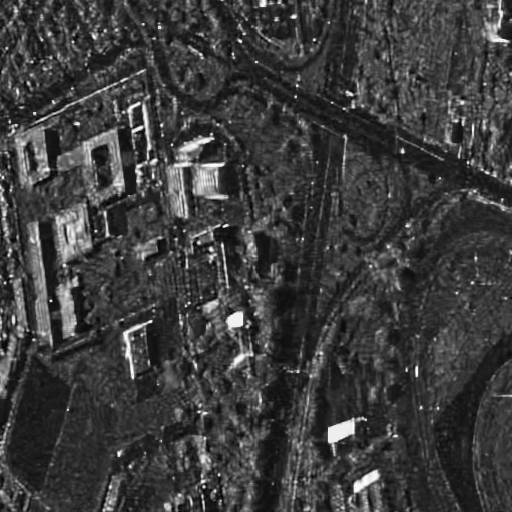} \label{s3}}
  \subfigure[]{\includegraphics[width=0.22 \linewidth]{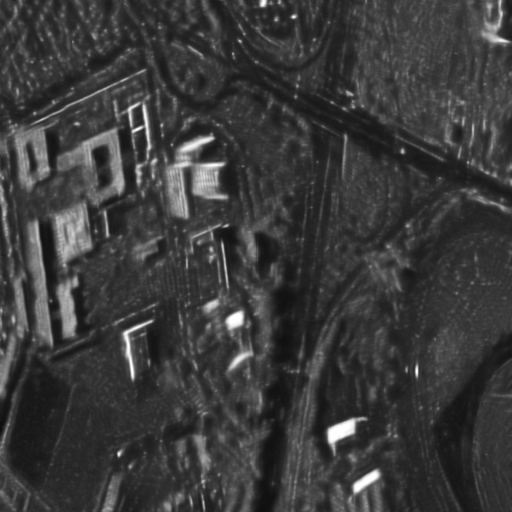} \label{s3}}
  \caption{Filtered images for real SAR.  (a) Frost. (b) PPB. (c) SAR-BM3D. (d) Proposed method.}\label{figure4}
\end{figure*}
\begin{table*}[htbp]
 \centering
 \caption{PSNR and SSIM Result for Barbara. The Best Results Are in Boldface}\label{1}
\begin{tabular}{|c|c|c|c|c|c|c|c|c|}
  \hline
  \hline
    & \multicolumn{4}{|c|}{PSNR} & \multicolumn{4}{|c|}{SSIM} \\
  \hline &$L=1$ &$L=2$ &$L=4$ &$L=8$ &$L=1$ &$L=2$ &$L=4$ &$L=8$ \\
  \hline
  \hline
  Frost & 17.71 & 19.87&21.73 & 23.24 & 0.39&0.38 & 0.47 & 0.56 \\
  PPB & 23.25 & 25.40& 27.58 & 29.57 & 0.63 & 0.71 & 0.80 & 0.86 \\
  SAR-BM3D & 25.41 & 27.44& 29.11 &30.72&0.73&0.81&0.86&0.89\\
  Proposed & $\mathbf{25.57}$ & $\mathbf{27.55}$& $\mathbf{29.53}$& $\mathbf{32.21}$& $\mathbf{0.74}$& $\mathbf{0.83}$& $\mathbf{0.89}$& $\mathbf{0.93}$\\
  \hline
  \hline
\end{tabular}
\end{table*}

\subsection{Parameter Setting}
We compare the proposed method with three common used denoising methods: Frost \cite{IEEE:two}, PPB \cite{IEEE:eleven}, the SAR-oriented version of block-matching 3-D (SAR-BM3D) \cite{IEEE:seven}. Such techniques have been selected because of their performance and availability of the codes. In all experiments, without explicit indication, the parameters of the aforementioned algorithms are set as suggested in the referenced papers. And our method is implemented by setting the image patch size to $7 \times 7$ pixels, the search area is restricted to an $81 \times 81$ neighborhood pixels, and the number of patches in a group $M=90$.
\subsection{Experimental Results}
For simulated SAR image, two objective criteria, namely peak signal-to-noise ratio (PSNR) and structure similarity index (SSIM) \cite{IEEE:twenty}, are adopted to provide quantitative quality evaluations of the denoising method. Table I reports the PSNR and SSIM results for Barbara.  We can see that the proposed method achieves the best performance on both PSNR and SSIM. Fig. 4 shows the denoised results of simulated speckle image. Frost (Fig. 4(c)) cannot suppress strong noise. PPB (Fig. 4(d)) is capable of reducing strong speckle noise. However it erases image details too much. As shown in Fig. 4(e) SAR-BM3D is able to preserve image details, however some strong speckle noise are not filtered out. From Fig. 4(f), we can see that the proposed method preserves image details meanwhile suppress strong noise.
\newline
\indent
The real SAR image used to evaluate the despeckling is TerraSAR-X sample image. Fig. 5 shows the filtered images. We can see that PPB and the proposed method have the best speckle-reduction ability in smooth areas, but PPB algorithm looks oversmoothing the sharp boundaries and faint details. In terms of detail preservation, SAR-BM3D and the proposed method achieve edges and faint details preservation. So, from the view of visual effect, we can conclude that the proposed method outperforms the other methods. To further evaluate performance, ENL is adopted to measure the speckle reduction in homogeneous areas (white rectangle box in Fig. 3(b)). Larger ENL values indicate stronger speckle rejection. The ENL values are calculated in table II.  From table II, we can see that ENL of proposed method is obviously higher than others, which indicates that the speckle-reduction abilities of our method outperform other three methods.

\section{Conclusion and Future Work}
In this letter, we have presented a novel SAR image despeckling method, which take advantages of the sparse decomposition and the nonlocal self-similarity techniques. The main idea of proposed principal dictionary denoising is to identify principal atoms to from an over-complete dictionary according to occurrence frequency of atoms over the similarity patches. Denoising based on such principal dictionary can effectively preserve detail information while suppressing strong noise.
\newline
\indent
Experiments conducted on both simulated and real SAR images demonstrated the effectiveness of the proposed method. The proposed method not only reduces the speckle noise but also preserves the image details. Future work includes the extensions of principal dictionary on a variety of applications, such as classification and PolSAR despeckling.
\section*{Acknowledgements}
 The idea of principal dictionary denoising presented here arises through a lot of deep discussions with Professor Henri Maitre at Telecom-ParisTech in France.


\begin{table}[htbp]
 \centering
 \caption{ENL of The Filtered Image. The Best Results Are in Boldface}\label{1}
\begin{tabular}{|c|c|c|c|c|c|}
  \hline
  \hline
   Method& Frost & PPB&SAR-BM3D & Proposed  \\
  \hline
  \hline
  ENL(Region 1)& 12.91& 80.2 & 34.4 & $\mathbf{86.4}$ \\
  ENL(Region 2)& 10.21& 76.2 & 28.1 & $\mathbf{80.3}$ \\
  \hline
  \hline
\end{tabular}
\end{table}

\ifCLASSOPTIONcaptionsoff
  \newpage
\fi

\end{document}